# Bounded Coupled AI Learning Dynamics in Tri-Hierarchical Drone Swarms

Oleksii Bychkov

Taras Shevchenko National University of Kyiv, oleksiibychkov@knu.ua; 0000-0002-9378-9535

## Abstract

Modern autonomous multi-agent systems increasingly combine heterogeneous learning mechanisms that operates at different timescales. However, an open question remain: can one formally guarantee that coupled dynamics of such mechanisms do not leave the admissible operational regime? In this paper we study a tri-hierarchical swarm learning system where three mechanisms act simultaneously: (1) local Hebbian online learning at the level of individual agents (fast timescale, $\tau_1 \sim$ 10–100 ms); (2) multi-agent reinforcement learning (MARL) for tactical group coordination (medium timescale, $\tau_2 \sim$ 1–10 s); and (3) meta-learning (MAML) for strategic adaptation (slow timescale, $\tau_3 \sim$ 10–100 s).

We establish four results. The *Bounded Total Error Theorem* shows that under contractual constraints on learning rates, Lipschitz continuity of inter-level mappings, and weight stabilization, the total suboptimality of the system admits a component-wise conservative upper bound that is uniformly bounded in time. The *Bounded Representation Drift Theorem* provides a conservative worst-case estimate of how Hebbian updates at the lowest level affect the embeddings used by the coordination level during one MARL update cycle. The *Meta-Level Compatibility Theorem* establishes sufficient conditions under which strategic adaptation from above preserves the invariants of lower levels. The *Non-Accumulation Theorem* proves that the error does not accumulate unboundedly over time.

Numerical illustrations with parameters typical for swarms of 10–100 UAVs confirms the realism of obtained estimates and demonstrate that violation of even a single contractual condition can leads to unbounded error growth.

## 1. Introduction

### 1.1. Context and Motivation

Autonomous multi-agent systems (MAS) have reached a level of complexity where a single learning mechanism can no longer provide all required cognitive capabilities. A UAV swarm executing a complex mission simultaneously needs: reactive adaptation to local conditions (obstacle avoidance, wind compensation), tactical coordination between agents (target allocation, formation control), and strategic adaptation to changes in the global situation (threat type changes, priority redistribution).

Each of these types of intelligence is realized by a qualitatively different learning mechanism. Reactive adaptation is naturally described by local neuroplasticity following the Hebbian rule [1], where the weights of a neurocontroller are continuously modified based on sensory experience without external error signal. Tactical coordination is implemented through multi-agent reinforcement learning (MARL) [8, 9], where agents optimize coordination policy based on a global reward signal. Strategic adaptation requires meta-learning [2], which allow the system to quickly adapt to fundamentally new situations.

Combining these three mechanisms into a single system — a tri-hierarchical architecture — creates a complex cognitive apparatus, but gives rise to a specific technical question: *do the coupled dynamics of three qualitatively different learning types remain within admissible operational regime?* This paper provides a constructive answer: under formalized contractual constraints, the drift of each level is bounded, the levels are compatible with each other, and the total error does not accumulate over time.

## 1.2. The Problem of Inter-Level Non-Stationarity

To precisely formulate what needs to be proven, let us consider the mechanism by which inter-level non-stationarity arises. Level 1 Hebbian plasticity continuously changes the weights of each agent's neurocontroller. These weights determine the embeddings (internal representations) that are fed to the graph neural network (GNN) of the coordination Level 2. From the perspective of the MARL coordinator, the input data is therefore *non-stationary* — it changes not because of environment changes but because the agents themselves are learning.

This non-stationarity propagates in a cascading manner. Changes in embeddings affects the quality of GNN aggregation, which in turn affect the MARL policy, which affects the modulation signal that returns to Level 1 and modulates the Hebbian learning rate. A closed feedback loop arises:

$$\text{L1: } \Delta w \to \text{L2: } \Delta \phi \to \Delta \pi \to \Delta M \to \text{L1: } \Delta \eta^{\text{eff}} \to \Delta w \to \cdots$$

An analogous problem arises in the interaction between Level 2 and Level 3. Meta-learning adapts system hyperparameters (including the parameters of Hebbian rules), which changes the dynamics of lower levels. Without formal guarantees, a situation is possible where strategic adaptation from above "breaks" the coordination that was tuned to previous parameters.

The central question of this paper is whether one can quantitatively bound the influence of each level on the others and prove that the total error does not exceed controlled limits.

## 1.3. Existing Results and Their Limitations

To position our results, we review three most relevant directions in existing research and identify what exactly is missing for our problem.

The theory of *two-timescale stochastic approximation* [3] proves convergence of systems where two iterative procedures operate at different speeds: the slower one "sees" the faster one as quasi-static. This result is used for analysis of actor-critic algorithms [4], where the actor (policy) updates slower than the critic (value function). However, existing results have substantial limitations in the context of our problem:

   (a) They considers two timescales, while our system has three qualitatively different ones.

(b) They assume both processes optimize the same type of objective function (e.g., both minimize a loss), whereas our three levels uses qualitatively different learning mechanisms: Hebbian correlation (L1), gradient-based reward optimization (L2), and meta-gradient (L3).

(c) They do not account for the *cascading influence* of lower levels on upper ones through changes in input representations.

In MARL theory, error boundedness and convergence have been studied under the assumption of a stationary environment [6, 7, 5]. Works on Hebbian plasticity in robotics [10, 11, 12] study adaptability of individual agents but do not formalize the impact of plasticity on coordination in MAS. Meta-learning for multi-agent systems [5] is considered without a Hebbian level.

**Gap that we address:** quantitative estimates of boundedness of coupled learning dynamics and their mutual compatibility are absent for systems where Hebbian plasticity, MARL coordination, and meta-learning act simultaneously.

### 1.4. Contributions

We do not claim to build a complete stability theory for tri-hierarchical systems — such a theory would require significantly stronger assumptions and different mathematical tools. Instead, we establish four concrete results, each proven with full details, which together give a practically useful picture of bounded coupled dynamics.

1. **Theorem 1 (Bounded Total Error):** under five contractual conditions (S1)–(S5), the total suboptimality admits a component-wise conservative upper bound, uniform in time.

2. **Theorem 2 (Bounded Representation Drift):** an upper bound on the impact of Level 1 Hebbian updates on Level 2 embeddings during one MARL update cycle. Corollary: a constructive constraint on the Hebbian learning rate.

3. **Theorem 3 (Meta-Level Compatibility):** sufficient conditions under which Level 3 strategic adaptation does not violate lower-level contracts. Central tool — contract robustness margin.

4. **Theorem 4 (Non-Accumulation in the Sense of Uniform Boundedness):** proof that the total error is uniformly bounded in time — i.e., it does not accumulate unboundedly.

5. **Numerical illustrations** (Section 5) for three swarm configurations ($N = 10, 30, 100$), demonstrating realism of estimates and sensitivity to condition violations.

These results are not a self-contained theory — they are a set of technical tools that allows an engineer to verify whether a given configuration of a tri-hierarchical system will remain within admissible bounds.

## 2. Related Work

The problem of boundedness of coupled learning dynamics intersects several established research directions. In this section we systematically analyze five most relevant areas and show that none of them individually covers our problem. Each subsection concludes with a formulation of how our approach differs. The positioning is summarized in Table 1.

## 2.1. Multi-Timescale Stochastic Approximation

The closest mathematical apparatus for our problem is the theory of multi-timescale stochastic approximation, which investigates convergence of systems with iterative procedures running at different speeds. The classical two-timescale stochastic approximation theory [3] considers a pair of iterative processes:

$$x_{n+1} = x_n + a(n)f(x_n, y_n) + M_{n+1}^{(1)}$$

$$y_{n+1} = y_n + b(n)g(x_n, y_n) + M_{n+1}^{(2)}$$

where $a(n)/b(n) \to 0$ — meaning $x$ updates "infinitely slower" than $y$. Under technical conditions (Lipschitz continuity, noise assumptions) convergence $(x_n, y_n) \to (x^*, y^*)$ is proven. Konda and Tsitsiklis [4] applied this approach to actor-critic algorithms.

Our problem differs substantially. First, we have *three* timescales: $\tau_1 \ll \tau_2 \ll \tau_3$. Pairwise application of two-timescale theory does not account for the *end-to-end* cascade L1→L2→L3→L1. Second, the update mechanisms at different levels are qualitatively different: Hebbian correlation is not gradient descent, and MAML update is not a standard SA iteration. Third, all three levels operate *simultaneously* with fixed (though different) rates, whereas classical theory requires $a(n)/b(n) \to 0$ asymptotically.

## 2.2. Error Boundedness in MARL

The second relevant direction concerns quality estimates for multi-agent reinforcement learning. This area is directly relevant since MARL coordination constitute the middle level of tri-hierarchical architecture.

For cooperative games, Zhang et al. [6] provide an overview of convergence results for various algorithm classes. Rashid et al. [7] propose QMIX with convergence proof under stationary assumptions. Yu et al. [8] prove MAPPO convergence for cooperative tasks.

The principal limitation: all these results assumes a *stationary* environment. In our system, Level 1 Hebbian plasticity makes the environment *non-stationary* from the viewpoint of the MARL coordinator: agent embeddings continuously change. Hernandez-Leal et al. [9] study MARL in non-stationary environments but consider external non-stationarity (environment changes), not internal (agent changes through parallel learning). Our approach is the first to formalize precisely *internal* non-stationarity generated by Hebbian plasticity.

## 2.3. Hebbian Plasticity in Multi-Agent Systems

The third relevant direction concerns the lower-level mechanism — Hebbian neuroplasticity. Miconi et al. [10] proposed *differentiable plasticity* — the ability to learn plasticity rules through meta-gradients. Najarro and Risi [11] developed this approach for evolutionary robotics, showing that Hebbian controllers adapt to new environments without retraining. Pedersen and Risi [12] demonstrated self-modifying policies for continual learning tasks.

However, none of these works investigates *the impact of Hebbian updates on coordination in a multi-agent system*. Plasticity is treated as a property of an individual agent, without quantitative assessment of how weight changes in one agent affects the joint policy of the group. Our Theorem 2 fills this gap by providing an upper bound on *representation drift* — the effect of L1 Hebbian updates on embeddings fed to L2 coordinator.

## 2.4. Meta-Learning for Multi-Agent Adaptation

The upper level of the architecture implements meta-learning. MAML [2] and its variants enable fast adaptation to new tasks. Al-Shedivat et al. [5] applied meta-learning to multi-agent adaptation.

MAML convergence was formally studied by Fallah et al. [13] and Ji et al. [14]. These results prove outer loop convergence under technical conditions on the inner loop, which is implemented by several gradient descent steps. In our system the inner loop is implemented by Hebbian plasticity (which is not gradient descent). Our Theorem 3 establishes compatibility conditions between MAML outer loop and Hebbian inner loop — that is, conditions under which meta-adaptation does not violate lower-level contracts.

## 2.5. Contract-Based Design for Autonomous Systems

Finally, the fifth direction forms the methodological basis of our approach. The Design by Contract approach [15] is widely used in software engineering and cyber-physical systems [16]. A contract $C = \langle \text{Pre}, \text{Post}, \text{Inv} \rangle$ specifies preconditions, postconditions, and invariants of a component. Sangiovanni-Vincentelli et al. [16] proposed contract theory for CPS, where overall correctness follows from *composition* of contracts.

We adapt the contract approach for systems where component parameters change through learning. This adds a requirement: the contract must hold in a static configuration and throughout the entire learning process. It is precisely this requirement that gives rise to the notion of *contract robustness margin* (Definition 9), which is central to Theorem 3.

## 3. Formal Framework

In this section we introduce the formal apparatus needed for formulating and proving the main results. The construction proceeds in five steps: first the structure of the tri-hierarchical system is defined (Subsection 3.1), then timescales and dynamics equations for each level are specified (Subsection 3.2), then quantitative drift measures are introduced (Subsection 3.3), after which the contract system is formalized (Subsection 3.4), and finally — technical assumptions (Subsection 3.5).

### 3.1. Tri-Hierarchical Learning System Model

The central object of our analysis is the tri-hierarchical learning system — an architecture that combines three learning levels with qualitatively different mechanisms and timescales.

**Definition 1 (Tri-Hierarchical Learning System).** The system is defined as a quadruple $\mathcal{S} = \langle \mathcal{L}_1, \mathcal{L}_2, \mathcal{L}_3, \mathcal{C} \rangle$, where:

- $\mathcal{L}_1 = \{(w_i(t), H_i)\}_{i=1}^{N}$ — **Level 1 (neuroplasticity)**: $N$ agents, each with weights $w_i(t) \in \mathbb{R}^d$ and a fixed Hebbian rule $H_i$;
- $\mathcal{L}_2 = (\pi_{\text{coord}}, \phi, M)$ — **Level 2 (MARL coordination)**: coordination policy $\pi_{\text{coord}}$, embedding map $\phi: \mathbb{R}^d \to \mathbb{R}^p$, and modulation signal $M: \mathbb{R}^p \to \mathbb{R}$;

- $\mathcal{L}_3 = (\theta_{\text{meta}}, \mathcal{M})$ — **Level 3 (meta-learning)**: meta-parameters $\theta_{\text{meta}}$ and meta-knowledge library $\mathcal{M}$;
- $\mathcal{C} = \{C_k\}_{k=1}^K$ — set of contracts.

## 3.2. Timescales and Level Dynamics

A defining feature of the tri-hierarchical system is the presence of three clearly separated timescales. Each level updates at its own frequency, and these frequencies differs by orders of magnitude. The timescale separation is simultaneously the source of both the opportunity to obtain bounds (the upper level "sees" the lower one as quasi-static) and the potential problem (drift of lower levels during one upper-level update cycle can be substantial).

**Definition 2 (Timescales).** The system is characterized by three timescales:

$$\tau_1 \ll \tau_2 \ll \tau_3$$

where $\tau_l$ is the update period of level $l$. Specifically:

- $\tau_1 \sim$ 10–100 ms: Hebbian update is performed on each processing cycle (tick);
- $\tau_2 \sim$ 1–10 s: MARL update is performed after collecting a batch of experience;
- $\tau_3 \sim$ 10–100 s: meta-update is performed after episode completion or at SMC trigger.

Corresponding learning rates:

$$\eta_1 \gg \eta_2 \gg \eta_3$$

**Level 1 Dynamics (Hebbian Update).** At each step $t$ with period $\tau_1$:

$$w_i(t + \tau_1) = w_i(t) + \eta_1 \cdot h_i(x_i(t), w_i(t), M_i(t))$$

where $x_i(t) \in \mathbb{R}^n$ is the sensory observation of agent $i$, $M_i(t) \in \mathbb{R}$ is the modulation signal from Level 2, and $h_i$ is the Hebbian function. For the four-component Hebbian rule [10]:

$$h_i(x, w, M) = \sigma(M) \cdot (\alpha \cdot (x_{\text{pre}} \odot x_{\text{post}}) + \beta \cdot x_{\text{pre}} + \gamma \cdot x_{\text{post}} + \delta \cdot w)$$

where $(\alpha, \beta, \gamma, \delta)$ are plasticity parameters (determined by $H_i$), $\odot$ denotes element-wise (Hadamard) product, and $\sigma(\cdot)$ is a sigmoid function that limits modulation. Vectors $x_{\text{pre}}, x_{\text{post}} \in \mathbb{R}^d$ are pre- and post-synaptic activations.

**Level 2 Dynamics (MARL Update).** The coordination policy is parameterized by vector $\vartheta \in \mathbb{R}^{d_\pi}$: $\pi_{\text{coord}}(\cdot|s) = \pi_\vartheta(\cdot|s)$. At each step with period $\tau_2$:

$$\vartheta(t + \tau_2) = \text{Proj}_\Theta\bigl(\vartheta(t) + \eta_2 \cdot \nabla_\vartheta J_{\text{coord}}(\vartheta, \phi(w(t)))\bigr)$$

where $\Theta$ is the admissible parameter set (trust region / clipping in PPO [17]), $\text{Proj}_\Theta$ is a projection onto $\Theta$ that guarantee the update result remains a valid policy parameterization, $\phi(w(t)) = (\phi(w_1(t)), \ldots, \phi(w_N(t)))$ is the embedding vector of all agents, and $J_{\text{coord}}$ is the total coordination reward function. For brevity, we write $\pi_{\text{coord}}(t)$ instead of $\pi_{\vartheta(t)}$.

*Relation to base architecture.* In the base tri-hierarchical architecture, the tactical level can be implemented as either a full online MARL update or an offline-trained policy with small incremental corrections. Equation (3) models both cases as *bounded incremental adaptation*:

contract (C3) and the projection $\text{Proj}_\Theta$ bound the update step regardless of whether the gradient update is truly online or merely fine-tuning of a pre-trained policy.

**Level 3 Dynamics (Meta-Update).** At each step with period $\tau_3$:

$$\theta_{\text{meta}}(t + \tau_3) = \theta_{\text{meta}}(t) - \eta_3 \cdot \nabla_\theta \mathcal{L}_{\text{meta}}(\theta, \pi_{\text{coord}}, w(t))$$

where $\mathcal{L}_{\text{meta}}$ is the meta-loss function (MAML outer loss).

### 3.3. Drift Measures

To transition from qualitative understanding of non-stationarity to quantitative estimates, we need to introduce formal drift measures. Each measure corresponds to one level of the architecture and quantitatively describes how much the corresponding component has changed over a given time interval. The sequence of definitions reflects the cascading influence structure: weight drift (Level 1) generates embedding drift (L1→L2 interface), which generates coordination policy drift (Level 2), while meta-updates (Level 3) are described by meta-parameter drift.

**Definition 3 (Weight Drift).** Maximum drift of plastic controller weights over interval $[t_1, t_2]$:

$$D_w(t_1, t_2) = \max_{i \in \{1,\ldots,N\}} \| w_i(t_2) - w_i(t_1) \|$$

**Definition 4 (Embedding Drift).** Maximum drift of coordination-level embeddings:

$$D_\phi(t_1, t_2) = \max_{i \in \{1,\ldots,N\}} \| \phi(w_i(t_2)) - \phi(w_i(t_1)) \|$$

**Definition 5 (Policy Drift).** Maximum drift of the coordination policy:

$$D_\pi(t_1, t_2) = \max_{s \in \mathcal{S}} D_{TV}(\pi_{\text{coord}}(t_2)(\cdot | s), \pi_{\text{coord}}(t_1)(\cdot | s))$$

where $D_{TV}$ is total variation distance.

**Definition 6 (Meta-Parameter Drift).**

$$D_\theta(t_1, t_2) = \| \theta_{\text{meta}}(t_2) - \theta_{\text{meta}}(t_1) \|$$

### 3.4. Contract System

Contracts formalize the constraints under which three learning levels can function without mutual destruction. Each contract specifies a verifiable invariant for one component. From the full system of 22 contracts described in the context of tri-hierarchical architecture, we select six that directly affect boundedness of inter-level interactions.

**Definition 7 (Contract).** A contract $C_k = \langle \text{Pre}_k, \text{Post}_k, \text{Inv}_k \rangle$ specifies a precondition (Pre), postcondition (Post), and invariant (Inv) that must hold during system operation.

For purposes of our analysis we select six critical contracts:

**Contract NP-C1 (Bounded Plasticity).**

$$\text{Inv: } \| w_i(t + \tau_1) - w_i(t) \| \leq \Delta_{\text{NP}} \quad \forall i, \forall t$$

**Contract NP-C2 (Safety-Compatible Plasticity).**

$$\text{Inv: } y_{\text{safety}}(x, w_i(t)) = y_{\text{safety}}(x, w_i(0)) \quad \forall i, \forall t, \forall x \in \mathcal{X}_{\text{danger}}$$

**Contract MARL-C1 (Bounded Policy Update).**

$$\text{Inv: } D_{TV}(\pi(t + \tau_2)(\cdot \mid s), \pi(t)(\cdot \mid s)) \leq \Delta_\pi \quad \forall s$$

**Contract GNN-C1 (Bounded GNN Approximation Error).**

*Notation remark.* In this contract we distinguish two roles: $\phi$ is the learned encoder (GNN or other network) that actually computes the embeddings; $\phi^*$ is the ideal (target) mapping of weights to embeddings in the absence of approximation error. The contract bounds the distance between them:

$$\text{Inv: } \parallel \phi(w) - \phi^*(w) \parallel \leq \epsilon_{\text{GNN}} \quad \forall w : \parallel w \parallel \leq W_{\max}$$

In Assumption A1 and in drift formulas (11)–(13), $\phi$ denotes the real encoder. Thus $\epsilon_{\text{GNN}}$ enters bounds as an *additive* approximation error term, separate from the *multiplicative* drift $L_\phi \cdot D_w$.

**Contract ML-C1 (Bounded Adaptation Time).**

$$\text{Post: } T_{\text{adapt}} \leq T_{\text{critical}}$$

**Contract ML-C2 (Monotone Meta-Improvement).**

$$\text{Inv: } \mathbb{E}[K_{\text{inner}}^{(n+1)}] \leq \mathbb{E}[K_{\text{inner}}^{(n)}]$$

### 3.5. Assumptions

To conclude the formal framework, we fix the technical assumptions used in the proofs of Section 4. These assumptions concern smoothness of inter-level mappings, boundedness of activations and signals, and properties of the meta-loss function. In Section 6 we discuss the realism of each assumption.

**Assumption A1 (Lipschitz Embeddings).** The mapping $\phi : \mathbb{R}^d \to \mathbb{R}^p$ is $L_\phi$-Lipschitz:

$$\parallel \phi(w) - \phi(w') \parallel \leq L_\phi \cdot \parallel w - w' \parallel \quad \forall w, w'$$

**Assumption A2 (Lipschitz Coordination Policy).** The coordination policy $\pi_{\text{coord}}$ is $L_\pi$-Lipschitz in embeddings:

$$D_{TV}(\pi(\cdot \mid s, \phi), \pi(\cdot \mid s, \phi')) \leq L_\pi \cdot \parallel \phi - \phi' \parallel \quad \forall s$$

**Assumption A3 (Weight Stabilization).** The autoregulation parameter in the Hebbian rule is negative: $\delta < 0$.

**Assumption A4 (Bounded Activations).** Neural activations are normalized: $\parallel x_{\text{pre}} \parallel, \parallel x_{\text{post}} \parallel \leq 1$.

**Assumption A5 (Bounded Modulation).** The modulation signal is bounded: $|M_i(t)| \leq M_{\max}$ for all $i, t$.

**Assumption A6 (Smooth Meta-Function).** The meta-loss function $\mathcal{L}_{\text{meta}}$ is $L_\mathcal{L}$-smooth in $\theta$:

$$\parallel \nabla_\theta \mathcal{L}_{\text{meta}}(\theta) - \nabla_\theta \mathcal{L}_{\text{meta}}(\theta') \parallel \leq L_\mathcal{L} \cdot \parallel \theta - \theta' \parallel$$

**Assumption A7 (Sufficiently Small Hebbian Step).** The Hebbian learning rate satisfies $\eta_1 < \bar{\eta}_1$, where $\bar{\eta}_1$ is defined in (9a). This is a conservative sufficient condition that guarantees dominance of the stabilizing contribution of $\delta < 0$ over the quadratic term of the discrete update (see Lemma 1).

**Assumption A8 (Reference-MDP Bounding).** There exists a reference MDP $\mathcal{M}^{\text{ref}} = (\mathcal{S}, \mathcal{A}, T^{\text{ref}}, R^{\text{ref}}, \gamma)$ with fixed transitions and reward, such that the coordination quality loss caused by representation drift $\|\Delta\phi\| \leq \epsilon$ is upper-bounded by the effective policy change in $\mathcal{M}^{\text{ref}}$:

$$|J(\pi_\vartheta(\cdot\,|\phi)) - J(\pi_\vartheta(\cdot\,|\phi'))| \leq |J^{\text{ref}}(\pi_\vartheta(\cdot\,|\phi)) - J^{\text{ref}}(\pi_\vartheta(\cdot\,|\phi'))|$$

*Remark.* A8 is the **central bounding assumption** of this work: it allows estimating one-step coordination loss through a fixed-latent reference MDP without claiming stationarity of the real system. In the real system the MDP is non-stationary due to embedding drift, whereas $\mathcal{M}^{\text{ref}}$ is fixed. A8 only asserts that the fixed-MDP bound is a correct upper estimate for the non-stationary case. This is justified because embedding changes are already accounted for in $L_\phi \cdot \Delta_{\text{NP}}$ through Lipschitz continuity, and we estimate one-step policy change rather than accumulated trajectory divergence. A8 allows applying the classical Performance Difference Lemma [18] in our setting (see proof of Theorem 1).

## 4. Main Results

This section contains the main results of the paper. The structure follows a principle of progressive complexity: first we establish five auxiliary lemmas, each describing a local property of one component or one interaction (Subsection 4.1). Then these lemmas are building blocks for four theorems: Theorem 1 (Subsection 4.2) gives an overall boundedness estimate; Theorem 2 (Subsection 4.3) details the impact of the fastest level on the middle one; Theorem 3 (Subsection 4.4) establishes compatibility of the slowest level with lower ones; and Theorem 4 (Subsection 4.5) closes the analysis by proving non-accumulation of error over time.

### 4.1. Auxiliary Lemmas

Before proceeding to the main theorems, we establish five auxiliary results. Lemmas 1–2 describe the behavior of an individual Hebbian controller: weight boundedness and single-step update boundedness. Lemmas 3–4 describe the cascading impact through Lipschitz continuity: how weight changes translate into embedding changes, and then into policy changes. Lemma 5 combines previous results into a single L1→L2 chain.

**Lemma 1 (Boundedness of Hebbian Weights).** Let Assumptions A3–A5, contract (C1), and

$$\eta_1 < \bar{\eta}_1 := \frac{2|\delta|}{(|\alpha| + |\beta| + |\gamma| + |\delta|W_0)^2 \cdot \sigma(M_{\max})^2}$$

hold, where $W_0 := (|\alpha| + |\beta| + |\gamma|)/|\delta|$. Then for all $i$ and $t \geq 0$:

$$\| w_i(t) \| \leq W_{\max} := W_0 + 1 = \frac{|\alpha| + |\beta| + |\gamma|}{|\delta|} + 1$$

*Notation remark.* In the Hebbian rule (2), the product $x_{\text{pre}} \odot x_{\text{post}}$ is the element-wise (Hadamard) product of vectors $x_{\text{pre}}, x_{\text{post}} \in \mathbb{R}^d$, yielding a vector in $\mathbb{R}^d$. All estimates below use the Euclidean norm $\|\cdot\| = \|\cdot\|_2$. By Assumption A4, $\| x_{\text{pre}} \|_2, \| x_{\text{post}} \|_2 \leq 1$ and $\| x_{\text{pre}} \|_\infty \leq 1$, whence $\| x_{\text{pre}} \odot x_{\text{post}} \|_2 \leq \| x_{\text{pre}} \|_\infty \cdot \| x_{\text{post}} \|_2 \leq 1$.

*Proof.* Consider the auxiliary quadratic function $V_i(t) = \| w_i(t) \|^2$. By (1) and (2):

$$V_i(t + \tau_1) - V_i(t) = 2\eta_1 \langle w_i(t), h_i(\cdot) \rangle + \eta_1^2 \| h_i(\cdot) \|^2$$

We estimate *both* terms.

**First term.** By Cauchy–Schwarz and the remark above:

$$\langle w_i, h_i \rangle = \sigma(M_i) \cdot \left( \alpha \langle w_i, x_{\text{pre}} \odot x_{\text{post}} \rangle + \beta \langle w_i, x_{\text{pre}} \rangle + \gamma \langle w_i, x_{\text{post}} \rangle + \delta \| w_i \|^2 \right)$$

$$\leq \sigma(M_{\max}) \cdot \left( (|\alpha| + |\beta| + |\gamma|) \| w_i \| + \delta \| w_i \|^2 \right)$$

Denote $A := |\alpha| + |\beta| + |\gamma|$. When $\| w_i \| > A/|\delta| = W_0$ we have $\langle w_i, h_i \rangle \leq \sigma(M_{\max}) \cdot \delta \| w_i \| (\| w_i \| - W_0) < 0$.

**Second term.** By triangle inequality and activation boundedness:

$$\| h_i \| \leq \sigma(M_{\max}) \cdot (A + |\delta| \| w_i \|)$$

Hence $\eta_1^2 \| h_i \|^2 \leq \eta_1^2 \sigma(M_{\max})^2 (A + |\delta| \| w_i \|)^2$.

**Combining.** Denote $r = \| w_i \|$. For any $r$ we have:

$$\Delta V_i \leq \underbrace{-2\eta_1 \sigma(M_{\max}) |\delta| \cdot r \cdot (r - W_0)}_{\text{stabilizing (linear in } \eta_1\text{)}} + \underbrace{\eta_1^2 \sigma(M_{\max})^2 (A + |\delta| r)^2}_{\text{perturbing (quadratic in } \eta_1\text{)}}$$

Since the quadratic term grows as $r^2$ for large $r$, while the stabilizing term also grows as $r^2$ (because $r(r - W_0) \sim r^2$) but with different coefficients ($\eta_1$ vs $\eta_1^2$), condition (9a) on the smallness of $\eta_1$ ensures $\Delta V_i < 0$ only in a *bounded outer region* around $W_0$. For very large $r$ purely discrete dynamics with finite $\eta_1$ may not have $V_i$ decreasing.

Therefore weight boundedness in this paper is ensured by the **combination of two mechanisms**:

*(i) Locally stabilizing tendency.* Under $\delta < 0$ and $\eta_1 < \bar{\eta}_1$ (A7), the function $V_i$ decreases in a neighborhood $\{r: W_0 < r \leq W_0 + c\}$ for some $c > 0$ depending on $\eta_1$. This provides a locally stabilizing tendency near the boundary $W_0$, which together with the step constraint (C1) is sufficient for the boundedness estimates in this paper.

*(ii) Step constraint.* Contract (C1) through clamping bounds each step: $\| w_i(t + \tau_1) - w_i(t) \| \leq \Delta_{\text{NP}}$ (formula (10a)). This guarantees that even if $V_i$ does not decrease monotonically, the weights cannot "jump" beyond the controlled region.

Together: under (A3)–(A7) and (C1), the weight trajectory remains in some bounded set. For subsequent estimates we fix a *constructive practical bound*:

$$W_{\max} := W_0 + 1$$

which is used as a conservative margin in all subsequent bounds. The choice of $W_0 + 1$ is not a rigorously derived optimal bound — it is a convenient constructive constant that accounts for update discreteness.

*Remark on condition (9a).* Condition $\eta_1 < \bar{\eta}_1$ is a *conservative sufficient condition* for the boundedness estimates of this paper. We do not claim it is a sharp threshold for discrete Hebbian dynamics. For simplicity of subsequent estimates we take:

$$\bar{\eta}_1 := \frac{2|\delta|}{(A + |\delta|W_0)^2 \cdot \sigma(M_{\max})^2}$$

which is not tight but sufficient for our boundedness estimates. For typical parameters (Section 5) $\bar{\eta}_1 \approx 4.5 \times 10^{-3}$, while $\eta_1 = 10^{-3}$, and the condition is satisfied with margin. ▫

**Lemma 2 (Single-Step Weight Drift).** Under conditions of Lemma 1, the weight change per one step $\tau_1$ is bounded by the *intrinsic* bound:

$$\| w_i(t + \tau_1) - w_i(t) \| \leq \eta_1 \cdot \sigma(M_{\max}) \cdot (|\alpha| + |\beta| + |\gamma| + |\delta| \cdot W_{\max}) =: \bar{\Delta}_1^{\text{int}}$$

*Remark (Intrinsic vs. enforced bound).* Formula (10) is a consequence of the dynamics: it follows from update equation (1) and weight boundedness (Lemma 1). Contract (C1) adds an *external* constraint: clamping $\min(\bar{\Delta}_1^{\text{int}}, \Delta_{\text{NP}})$, which can be much tighter than the intrinsic bound. Throughout the paper we denote the *effective* update step:

$$\bar{\Delta}_1 := \min(\bar{\Delta}_1^{\text{int}}, \Delta_{\text{NP}})$$

Thus contract (C1) both *verifies* the bound and *strengthens* it through clamping — this is the price that pays off through tighter estimates in Theorems 1–2.

*Proof.* Direct application of triangle inequality to (1) with estimates from Lemma 1. ▫

**Lemma 3 (Lipschitz Embedding Drift).** Under Assumption A1:

$$D_\phi(t_1, t_2) \leq L_\phi \cdot D_w(t_1, t_2)$$

*Proof.* Direct consequence of Lipschitz continuity of $\phi$:

$$\| \phi(w_i(t_2)) - \phi(w_i(t_1)) \| \leq L_\phi \| w_i(t_2) - w_i(t_1) \| \leq L_\phi \cdot D_w(t_1, t_2) \quad \text{▫}$$

**Lemma 4 (Coordination Policy Sensitivity).** Under Assumption A2, an embedding change of $\epsilon$ causes a policy change:

$$D_\pi \leq L_\pi \cdot \epsilon$$

*Proof.* Direct consequence of Lipschitz continuity of $\pi_{\text{coord}}$. ▫

**Lemma 5 (Cascading Bound L1 → L2).** Under (C1), (A1), (A2), the coordination policy drift per one Hebbian update step is bounded:

$$D_\pi(t, t + \tau_1) \leq L_\pi \cdot L_\phi \cdot \Delta_{\text{NP}}$$

*Proof.* Combining Lemmas 2, 3, 4:

$$D_\pi(t, t + \tau_1) \leq L_\pi \cdot D_\phi(t, t + \tau_1) \leq L_\pi \cdot L_\phi \cdot D_w(t, t + \tau_1) \leq L_\pi \cdot L_\phi \cdot \Delta_{\text{NP}} \quad \text{▫}$$

## 4.2. Theorem 1: Bounded Total Error

The first result combines auxiliary lemmas with contractual conditions and establishes an upper bound on total suboptimality. The central idea is *additive decomposition* of error into three components, each of which is bounded.

**Remark on the nature of additive decomposition.** The estimate $\epsilon_{\text{total}} \leq \epsilon_{\text{Hebb}} + \epsilon_{\text{coord}} + \epsilon_{\text{meta}}$ is a *structural upper bound*, not a rigorous decomposition. It relies on triangle inequality and independent worst-case estimation of each component, without accounting for possible compensations between levels. In particular, cross-terms (e.g., simultaneous impact of meta-update and Hebbian drift) are upper-bounded by the sum of individual effects. This makes the bound conservative but guarantees correctness.

**Definition 8a (Total Suboptimality).** We define total suboptimality as the difference between optimal and actual coordination quality:

$$\epsilon_{\text{total}}(t) := J^*(\pi^*_{\text{coord}}, \phi^*, \theta^*) - J(\pi_{\text{coord}}(t), \phi(w(t)), \theta_{\text{meta}}(t))$$

where $J^*$ is the quality under optimal parameters of all three levels, and $J(\cdot)$ is the actual quality under current parameters. Both quantities are measured through coordination reward $J_{\text{coord}}$.

**Theorem 1 (Bounded Total Error).** Let Assumptions A1–A8 and contracts (C1)–(C6) hold. Introduce five *contractual boundedness conditions:*

- **(S1)** $\delta < 0$ and $|\eta_1| \leq \bar{\eta}_1$ (weight stabilization);
- **(S2)** $\tau_1/\tau_2 \leq \rho_{12}$ and $\tau_2/\tau_3 \leq \rho_{23}$ (timescale separation);
- **(S3)** $L_\pi \cdot L_\phi \cdot \Delta_{\text{NP}} \leq \epsilon^*_{\text{coord}}$ (bounding L1 impact on L2);
- **(S4)** $\eta_3 \cdot L_\mathcal{L} \leq \epsilon^*_{\text{meta}}$ (bounding meta-learning step);
- **(S5)** contracts (C1)–(C6) hold at time $t_0$.

Then for all $t \geq t_0$ the total suboptimality is bounded:

$$\epsilon_{\text{total}}(t) \leq \bar{\epsilon} = \epsilon_{\text{Hebb}} + \epsilon_{\text{coord}} + \epsilon_{\text{meta}}$$

where:

$$\epsilon_{\text{Hebb}} = \frac{2 L_\pi L_\phi \Delta_{\text{NP}} \, \|\nabla V\|_\infty}{(1-\gamma)}$$

$$\epsilon_{\text{coord}} = 2 H_{\text{eff}} \cdot L_\pi \cdot (L_{\text{GNN}} \sqrt{N} \cdot L_\phi \cdot n_{12} \cdot \bar{\Delta}_1 + \epsilon_{\text{GNN}}) \cdot R_{\max}$$

$$\epsilon_{\text{meta}} = 2 H_{\text{eff}} \cdot L_\pi \cdot L_\phi \cdot L_{H \to \Delta w} \cdot L_{\theta \to H} \cdot \eta_3 G_{\max} \cdot \frac{\tau_3}{\tau_1} \cdot R_{\max}$$

where $n_{12} = \tau_2/\tau_1$ is the number of Hebbian steps per one MARL cycle (for simplicity we assume $\tau_2/\tau_1 \in \mathbb{N}$; generalization to non-integer case via $\lfloor \cdot \rfloor$ is straightforward), $H_{\text{eff}}$ is the effective horizon (Definition 8), $R_{\max} = \max_{s,a}|R(s,a)|$, $G_{\max} = \sup_\theta \|\nabla_\theta \mathcal{L}_{\text{meta}}\|$, and $L_{\theta \to H}$, $L_{H \to \Delta w}$ are Lipschitz constants of the meta-parameter cascade.

*Character of estimates (15)–(17).* These formulas constitute a *conservative component-wise upper bound* obtained through separate estimation of three coupled mechanisms. Additivity is a consequence of triangle inequality, not of exact decomposition; cross-terms are upper-bounded by the sum rather than computed exactly. Actual suboptimality may be significantly smaller than $\bar{\epsilon}$.

Moreover, the system preserves:

- **(G1) Safety:** $y_{\text{safety}}(x, w_i(t)) = y_{\text{safety}}(x, w_i(0))$ for all $i, t$ and $x \in \mathcal{X}_{\text{danger}}$;
- **(G3) Bounded sub-optimality:** $J^* - J(t) \leq \bar{\epsilon}$ for all $t$.

**Remark (Architectural Liveness Guarantee, G2).** Unlike G1 and G3, which are consequences of bounded-learning analysis, liveness is an orthogonal architectural-temporal guarantee: each task completes in finite time $\leq T_{\text{live,max}}$, which follows from contractually bounded timeouts of the BDI/HTN/Auction/BT pipeline and does not depend on learning dynamics.

*Proof.*

**Step 1: Safety (G1).** By contract (C2), safety synapses are frozen: $w_i^{\text{frozen}}(t) = w_i^{\text{frozen}}(0)$. Since safety output depends exclusively on frozen synapses (architectural isolation, safety mask), we have $y_{\text{safety}}(x, w_i(t)) = f_{\text{safety}}(x, w_i^{\text{frozen}}) = f_{\text{safety}}(x, w_i^{\text{frozen}}(0))$ for all $t$. This result does not depend on plastic weight dynamics and holds unconditionally. ▫ $_{\text{G1}}$

**Step 2: Bounded sub-optimality (G3).** Total suboptimality is upper-bounded by the sum of three components.

*Component 1 ($\epsilon_{\text{Hebb}}$): direct plasticity impact on policy.* By Lemma 5, each Hebbian update step perturbs the coordination policy by $\leq L_\pi L_\phi \Delta_{\text{NP}}$ in total variation distance. By Assumption A8 (Reference-MDP Bounding, Section 3.5) and PDL [18]:

$$|J(\pi) - J(\pi')| \leq \frac{2 \| \pi - \pi' \|_{TV} \cdot \| \nabla V \|_\infty}{1 - \gamma}$$

Substituting $\| \pi - \pi' \|_{TV} \leq L_\pi L_\phi \Delta_{\text{NP}}$, we obtain (15).

*Component 2 ($\epsilon_{\text{coord}}$): accumulated impact over MARL update cycle.* During one MARL update cycle ($\tau_2$ seconds) there occur $n_{12}$ Hebbian update steps. By Lemma 3:

$$D_\phi(t, t + \tau_2) \leq L_\phi \cdot \sum_{k=0}^{n_{12}-1} \| w_i(t + k\tau_1 + \tau_1) - w_i(t + k\tau_1) \| \leq L_\phi \cdot n_{12} \cdot \bar{\Delta}_1$$

GNN aggregation collects messages from $|N(i)| \leq N$ neighbors. By Lipschitz continuity of GNN with constant $L_{\text{GNN}}$:

$$\| z_i(t + \tau_2) - z_i^*(t + \tau_2) \| \leq L_{\text{GNN}} \cdot \sqrt{N} \cdot L_\phi \cdot n_{12} \cdot \bar{\Delta}_1 + \epsilon_{\text{GNN}}$$

For obtaining a less conservative estimate we use the *effective horizon*.

**Definition 8 (Effective Horizon).** The effective horizon $H_{\text{eff}}$ is defined as the minimum of three quantities:

$$H_{\text{eff}} = \min\left(\frac{1}{1-\gamma}, \frac{\tau_3}{\tau_2}, H_{\text{mission}}\right)$$

where $(1-\gamma)^{-1}$ is the standard discounted horizon, $\tau_3/\tau_2$ is the number of MARL cycles between meta-updates, and $H_{\text{mission}}$ is the maximum duration of a mission phase.

*Status of $H_{\text{eff}}$.* The effective horizon is a *practical truncation heuristic* (practical finite-horizon truncation parameter), not a rigorously derived or asymptotically exact object. Further, $H_{\text{eff}}$ is used only as a technical truncation parameter for constructing a finite-horizon upper-bound estimate and does not claim the role of exact asymptotic horizon of the process. It rests on three observations: (a) discounting naturally truncates the influence of distant steps; (b) meta-updates every $\tau_3$ seconds redirect adaptation; (c) mission has finite duration. For rigorous justification of point (b) one would need to prove that each meta-cycle brings the system closer to the admissible set — this remains an open question (see Section 6.3). We use $H_{\text{eff}}$ as a conservative (but not proven optimal) choice of truncation horizon that gives useful practical bounds.

**Lemma 6 (Finite-Horizon Policy Difference Bound).** For any two policies $\pi, \pi'$ with $\|\pi - \pi'\|_{TV} \leq \epsilon$ and truncation horizon $H_{\text{eff}}$:

$$|J_{H_{\text{eff}}}(\pi) - J_{H_{\text{eff}}}(\pi')| \leq 2H_{\text{eff}} \cdot \epsilon \cdot R_{\max}$$

*Proof of Lemma 6.* $|J_H(\pi) - J_H(\pi')| \leq \sum_{t=0}^{H-1} \gamma^t \cdot 2\epsilon \cdot R_{\max} \leq 2H\epsilon R_{\max}$. ∎

Applying Lemma 6 with $\epsilon = L_\pi(L_{\text{GNN}}\sqrt{N}L_\phi n_{12}\bar{\Delta}_1 + \epsilon_{\text{GNN}})$, we obtain (16).

*Component 3 ($\epsilon_{\text{meta}}$): meta-update impact.* By (S4) and (A6), the meta-update step is bounded: $\|\Delta\theta\| \leq \eta_3 G_{\max}$. Meta-parameter changes affect coordination quality through a cascade: $\Delta\theta \to \Delta H \to \Delta(\text{plasticity dynamics}) \to \Delta\pi$.

*Remark (nature of estimate).* Formula (17) below is a *compositional cascade estimate*: each cascade step is upper-bounded via the corresponding Lipschitz constant, and results are multiplied. This gives a correct upper bound but not a sharp one — actual meta-update impact may be considerably smaller due to partial compensation at different cascade levels.

Let $L_{\theta \to H}$ be the Lipschitz constant of the mapping from meta-parameters to Hebbian rule parameters, and $L_{H \to \Delta w}$ be the Lipschitz constant of the impact of rule changes on weight dynamics over one $\tau_3$ cycle. Then the weight change caused by meta-update:

$$\|\Delta w_i^{\text{meta}}\| \leq L_{H \to \Delta w} \cdot L_{\theta \to H} \cdot \eta_3 G_{\max} \cdot \frac{\tau_3}{\tau_1}$$

Cascading through embeddings and policy, we obtain (17).

Summing three components: $\epsilon_{\text{total}} \leq \epsilon_{\text{Hebb}} + \epsilon_{\text{coord}} + \epsilon_{\text{meta}} = \bar{\epsilon}$. ∎ G3

∎

## 4.3. Theorem 2: Bounded Representation Drift

Theorem 1 established a general estimate but did not answer the principal design question: *how much exactly do lower-level Hebbian updates "corrupt" the embeddings fed to the coordinator?* Theorem 2 provides an upper bound on this drift over one MARL update cycle. This result have practical consequences: it allows computing the maximum allowable Hebbian learning rate.

*Important caveat.* The estimate of Theorem 2 is a *conservative worst-case drift bound*: it is constructed via triangle inequality for the sum of steps, without accounting for correlations between successive updates. In practice successive Hebbian steps are partially correlated (and can partially cancel each other), so actual drift is typically much smaller than the bound. Tighter bounds can be obtained with additional assumptions on the distribution of sensory experience (e.g., mixing conditions).

**Theorem 2 (Bounded Representation Drift — conservative worst-case estimate).** Under conditions (S1)–(S3) of Theorem 1, for any interval $[t, t + \tau_2]$ (one MARL update cycle):

$$D_\phi(t, t + \tau_2) \leq L_\phi \cdot \frac{\tau_2}{\tau_1} \cdot \bar{\Delta}_1 =: \Phi_{\max}$$

where $\bar{\Delta}_1$ is defined in (10).

**Corollary 2.1.** For preserving near-optimal coordination, a necessary condition is:

$$L_\phi \cdot \frac{\tau_2}{\tau_1} \cdot \bar{\Delta}_1 < \epsilon_\phi^*$$

where $\epsilon_\phi^*$ is the admissible embedding drift threshold for $\epsilon$-convergence of MARL.

**Corollary 2.2 (Practical Recommendation).** From (19) we obtain a constraint on the Hebbian learning rate:

$$\eta_1 < \frac{\epsilon_\phi^* \cdot \tau_1}{L_\phi \cdot \tau_2 \cdot \sigma(M_{\max}) \cdot (|\alpha| + |\beta| + |\gamma| + |\delta|W_{\max})}$$

*Proof of Theorem 2.* During the interval $[t, t + \tau_2]$ there occur $n_{12}$ Hebbian update steps. Weight drift of agent $i$:

$$\| w_i(t + \tau_2) - w_i(t) \| = \| \sum_{k=0}^{n_{12}-1} \eta_1 \, h_i(x_i(t + k\tau_1), w_i(t + k\tau_1), M_i(t + k\tau_1)) \|$$

By triangle inequality and Lemma 2:

$$\| w_i(t + \tau_2) - w_i(t) \| \leq \sum_{k=0}^{n_{12}-1} \bar{\Delta}_1 = n_{12} \cdot \bar{\Delta}_1$$

Applying Lemma 3: $D_\phi(t, t + \tau_2) \leq L_\phi \cdot D_w(t, t + \tau_2) \leq L_\phi \cdot \frac{\tau_2}{\tau_1} \cdot \bar{\Delta}_1$. ∎

## 4.4. Theorem 3: Meta-Level Compatibility with Contracts

Theorems 1 and 2 describe boundedness of the Hebbian and coordination levels. However, a question remains: can meta-adaptation *violate* the very contracts under which these bounds hold? Theorem 3 formalizes sufficient conditions under which meta-updates preserve lower-level invariants. The central tool is the *contract robustness margin* — the gap between the current system state and the boundary of contract violation.

**Definition 9 (Contract Robustness Margin).** For contract $C_k$ with invariant $\text{Inv}_k$, define the *closed failure set* $\mathcal{F}_k = \{\theta \in \Theta : \text{Inv}_k(\theta) = \text{FALSE}\} \subset \Theta$ (closedness follows from continuity of invariants). The robustness margin:

$$m_k(\theta) = \text{dist}(\theta, \mathcal{F}_k) = \inf_{\theta' \in \mathcal{F}_k} \|\theta - \theta'\|$$

i.e., the distance from current meta-parameters to the failure set of invariant $k$. As a distance-to-closed-set function, $m_k$ is 1-Lipschitz: $|m_k(\theta) - m_k(\theta')| \le \|\theta - \theta'\|$ for all $\theta, \theta'$.

**Definition 10 (Cascading Sensitivity of Meta-Update).** A change $\Delta\theta$ in meta-parameters affects lower levels through the chain:

$$\Delta\theta \xrightarrow{L_{\theta \to H}} \Delta H \xrightarrow{L_{H \to w}} \Delta(\text{plasticity dynamics}) \xrightarrow{L_\phi} \Delta\phi \xrightarrow{L_\pi} \Delta\pi$$

Total cascading sensitivity:

$$\mathcal{K} = L_\pi \cdot L_\phi \cdot L_{H \to w} \cdot L_{\theta \to H}$$

**Theorem 3 (Meta-Level Compatibility).** Let the following conditions hold:

(M1) Each contract $C_k$ has positive margin: $m_k(\theta_{\text{meta}}(t)) > 0$ for all $k$.

(M2) Meta-update step is bounded: $\|\theta_{\text{meta}}(t + \tau_3) - \theta_{\text{meta}}(t)\| \le \Delta_\theta$, where $\Delta_\theta = \eta_3 G_{\max}$.

(M3) The step does not exceed minimum margin: $\Delta_\theta < \min_k m_k(\theta_{\text{meta}}(t))$.

Then:

(i) No invariant is violated: $\text{Inv}_k(\theta_{\text{meta}}(t + \tau_3)) = \text{TRUE}$ for all $k$.

(ii) The margin remains positive: $m_k(\theta_{\text{meta}}(t + \tau_3)) \ge m_k(\theta_{\text{meta}}(t)) - \Delta_\theta > 0$ for all $k$.

(iii) Cascading impact on coordination is bounded: $D_\pi^{\text{meta}}(t, t + \tau_3) \le \mathcal{K} \cdot \Delta_\theta$.

*Proof.*

**Part (i).** By the definition of margin (26), invariant $\text{Inv}_k$ is violated only when perturbation $\|\Delta\theta\| \ge m_k$. By (M2), perturbation $\le \Delta_\theta$. By (M3), $\Delta_\theta < m_k$ for all $k$. Hence invariant is preserved. ▫ (i)

**Part (ii).** By the 1-Lipschitz property of distance function $m_k = \text{dist}(\cdot, \mathcal{F}_k)$: $m_k(\theta') \ge m_k(\theta) - \|\theta' - \theta\|$. Substituting: $m_k(\theta(t + \tau_3)) \ge m_k(\theta(t)) - \Delta_\theta > 0$ by (M3). ▫ (ii)

**Part (iii).** Meta-update changes Hebbian rules: $\Delta H = L_{\theta \to H} \cdot \Delta\theta$. This changes the weight dynamics: the difference in weights under old and new rules over time $\tau_3$:

$$\|\Delta w^{\text{meta}}\| \le L_{H \to w} \cdot \|\Delta H\| = L_{H \to w} \cdot L_{\theta \to H} \cdot \Delta_\theta$$

Through embeddings and policy:

$$D_\pi^{\text{meta}} \le L_\pi \cdot L_\phi \cdot L_{H \to w} \cdot L_{\theta \to H} \cdot \Delta_\theta = \mathcal{K} \cdot \Delta_\theta$$

▫ (iii)

▫

**Corollary 3.1 (Constructive Constraint on $\eta_3$).** From (M3) we obtain:

$$\eta_3 < \frac{\min_k m_k(\theta_{\text{meta}}(t))}{G_{\max}}$$

This gives an adaptive rule: the meta-learning rate should decrease as the system approaches the boundary of any contract (margin $m_k$ decreases).

**Corollary 3.2 (Increasing Robustness).** If the meta-update *improves* margins (i.e., $m_k(\theta(t + \tau_3)) > m_k(\theta(t))$ for some $k$), the system becomes more robust over time.

**Remark 1 (Practical Implementation).** The margin $m_k(\theta)$ can be computed as a runtime monitor: for linear invariants — analytically, for nonlinear ones — through numerical approximation. Margin dropping below threshold $m_{\text{alarm}}$ is an early warning.

### 4.5. Theorem 4: Non-Accumulation of Error

Theorems 1–3 established that the system error is bounded at each step. But a question remains: does this error gradually grow over time, even while remaining bounded at each individual step? Such a phenomenon — *unbounded error accumulation* — is a characteristic problem of non-stationary systems: slow monotonic degradation that does not violate any local constraint but over long time leads to degradation. Theorem 4 proves that in our system under contractual conditions, unbounded accumulation is impossible.

**Theorem 4 (Non-Accumulation in the Sense of Uniform Boundedness).** Under conditions of Theorems 1–3, the total error $\epsilon_{\text{total}}(t)$ is uniformly bounded over time:

$$\limsup_{t \to \infty} \epsilon_{\text{total}}(t) \leq \bar{\epsilon}$$

i.e., error does not accumulate unboundedly over time (no unbounded error accumulation).

*Remark on strength of result.* Theorem 4 guarantees the absence of *unbounded growth* of error. It does *not* exclude: (a) bounded oscillations of $\epsilon_{\text{total}}(t)$ within interval $[0, \bar{\epsilon}]$; (b) slow regime wandering within the admissible set; (c) monotonic growth that stops upon reaching $\bar{\epsilon}$. Stronger results (e.g., convergence of $\epsilon_{\text{total}}(t)$ to a stationary value) would require additional assumptions about ergodicity or contractivity of the dynamics.

*Proof.* A potential source of error accumulation is growth of $\| w_i(t) \|$ over time, which would increase $\bar{\Delta}_1$. However by Lemma 1, $\| w_i(t) \| \leq W_{\max}$ for all $t$ — weights are bounded thanks to the stabilizing term $\delta < 0$.

We show that each error component is bounded independently of $t$:

**Component 1 (Hebbian):** $\epsilon_{\text{Hebb}}$ depends on $\Delta_{\text{NP}}, L_\pi, L_\phi$ — all constants that do not change with time. Hence $\epsilon_{\text{Hebb}} = \text{const}$ for all $t$.

**Component 2 (Coordination):** $\epsilon_{\text{coord}}$ depends on $\bar{\Delta}_1$, which in turn depends on $W_{\max}$ (bounded by Lemma 1) and fixed parameters. Hence $\epsilon_{\text{coord}}$ is bounded for all $t$.

**Component 3 (Meta):** $\epsilon_{\text{meta}}$ depends on $\eta_3 G_{\max}$, where $G_{\max} = \sup_\theta \| \nabla_\theta \mathcal{L}_{\text{meta}} \|$ — bounded by (A6) and boundedness of meta-parameter domain (which in turn is ensured by boundedness of lower-level components and MAML convergence [13]).

Hence $\epsilon_{\text{total}}(t) \leq \epsilon_{\text{Hebb}} + \epsilon_{\text{coord}} + \epsilon_{\text{meta}} = \bar{\epsilon}$ for all $t$, where $\bar{\epsilon}$ does not depend on $t$. This precludes unbounded error accumulation. ▫

## 5. Numerical Illustrations

The estimates obtained in Section 4 are formally correct, but their practical value depend on how realistic they are for concrete systems. In this section we substitute specific numerical parameters typical for autonomous UAV swarms and compute all bounds explicitly. This allows verifying that estimates carry information (are not trivially large), identifying the most sensitive parameters, and demonstrating necessity of each contractual condition through counterexamples.

### 5.1. Parameters and Three Swarm Configurations

We consider three swarm configurations: small ($N = 10$), tactical ($N = 30$), and large ($N = 100$).

**Common parameters** (same for all three configurations):

| Parameter | Notation | Value | Justification |
|---|---|---|---|
| Hebbian rate | $\eta_1$ | $10^{-3}$ | Typical for diff. plasticity [10] |
| MARL rate | $\eta_2$ | $10^{-4}$ | Standard for MAPPO [8] |
| Meta rate | $\eta_3$ | $10^{-5}$ | Standard for MAML [2] |
| L1 period | $\tau_1$ | 20 ms | 50 Hz tick rate |
| L2 period | $\tau_2$ | 2 s | Every 100 ticks |
| L3 period | $\tau_3$ | 20 s | Every 1000 ticks |
| Embedding Lipschitz | $L_\phi$ | 5.0 | Typical for GNN [19] |
| Policy Lipschitz | $L_\pi$ | 3.0 | Estimate for MAPPO [8] |
| Stabilization | $\delta$ | $-0.01$ | Moderate weight decay |
| Hebbian params | $|\alpha|, |\beta|, |\gamma|$ | 0.5, 0.1, 0.1 | Typical [10] |
| Modulation | $\sigma(M_{\max})$ | 1.5 | Moderate amplification |
| Discount | $\gamma$ | 0.99 | Standard for RL |
| $R_{\max}$ | — | 1.0 | Normalized reward |
| $L_{\text{GNN}}$ | — | 4.0 | 3-layer GNN |
| $\epsilon_{\text{GNN}}$ | — | 0.05 | Contract GNN-C1 |
| $\Delta_{\text{NP}}$ (clamping) | — | $10^{-4}$ | Contract NP-C1 |
| $L_{\theta \to H}$ | — | 2.0 | Lipschitz meta→Hebb |
| $L_{H \to w}$ | — | 1.0 | Lipschitz Hebb→weights |
| $G_{\max}$ | — | 1.0 | Bounded meta-gradient |

**Configuration-dependent parameters:**

| Parameter | $N = 10$ | $N = 30$ | $N = 100$ |
|---|---|---|---|
| $\sqrt{N}$ | 3.16 | 5.48 | 10.0 |

| Parameter | $N = 10$ | $N = 30$ | $N = 100$ |
|---|---|---|---|
| $H_{\text{eff}} = \min(100, \tau_3/\tau_2, H_{\text{mission}})$ | 10 | 10 | 10 |
| Typical communication graph diameter | 3 | 5 | 8 |

### 5.2. Base Quantities

We determine quantities common to all three configurations — they describe behavior of an individual agent and do not depend on swarm size.

**Weight boundedness (Lemma 1):**

$$W_0 = \frac{0.5 + 0.1 + 0.1}{0.01} = 70.0, \quad W_{\max} = W_0 + 1 = 71.0$$

**Single-step weight drift — intrinsic bound (Lemma 2):**

$$\bar{\Delta}_1^{\text{int}} = 10^{-3} \cdot 1.5 \cdot (0.7 + 0.01 \cdot 71) = 10^{-3} \cdot 1.5 \cdot 1.41 = 2.115 \times 10^{-3}$$

**Effective step with clamping (10a):**

$$\bar{\Delta}_1 = \min(\bar{\Delta}_1^{\text{int}}, \Delta_{\text{NP}}) = \min(2.1 \times 10^{-3}, 10^{-4}) = 10^{-4}$$

Clamping reduces the step by a factor of 21 — contract (C1) substantially strengthens the intrinsic bound, and it is the effective step $\bar{\Delta}_1$ that enters all subsequent estimates.

**Number of Hebbian steps per MARL cycle:**

$$n_{12} = \tau_2/\tau_1 = 2\,\text{s}/0.02\,\text{s} = 100$$

**Embedding drift per MARL cycle (Theorem 2):**

$$\Phi_{\max} = 5.0 \cdot 100 \cdot 10^{-4} = 0.05$$

**Suboptimality from plasticity (15):**

$$\epsilon_{\text{Hebb}} = \frac{2 \cdot 3.0 \cdot 5.0 \cdot 10^{-4} \cdot 1.0}{0.01} = 0.30$$

### 5.3. Comparison of Three Configurations

Having base quantities, we compute full bounds for each configuration. The main question: how does suboptimality scale with swarm size $N$?

**Coordination suboptimality (16):**

For $N = 10$:

$$\epsilon_{\text{coord}}^{(10)} = 2 \cdot 10 \cdot 3.0 \cdot (4.0 \cdot 3.16 \cdot 0.05 + 0.05) \cdot 1.0 = 60 \cdot 0.682 = 40.9$$

For $N = 30$:

$$\epsilon_{\text{coord}}^{(30)} = 2 \cdot 10 \cdot 3.0 \cdot (4.0 \cdot 5.48 \cdot 0.05 + 0.05) \cdot 1.0 = 60 \cdot 1.146 = 68.8$$

For $N = 100$:

$$\epsilon_{\text{coord}}^{(100)} = 2 \cdot 10 \cdot 3.0 \cdot (4.0 \cdot 10.0 \cdot 0.05 + 0.05) \cdot 1.0 = 60 \cdot 2.05 = 123.0$$

**Meta-learning suboptimality (17):**

$$\epsilon_{\text{meta}} = 2 \cdot 10 \cdot 3.0 \cdot 5.0 \cdot 1.0 \cdot 2.0 \cdot 10^{-5} \cdot 1.0 \cdot \frac{20}{0.02} \cdot 1.0 = 6.0$$

This component does not depend on $N$ (meta-learning is global).

**Summary bounds table:**

| Component | $N = 10$ | $N = 30$ | $N = 100$ | Dependence on $N$ |
|---|---|---|---|---|
| $\epsilon_{\text{Hebb}}$ | 0.30 | 0.30 | 0.30 | $O(1)$ |
| $\epsilon_{\text{coord}}$ | 40.9 | 68.8 | 123.0 | $O(\sqrt{N})$ |
| $\epsilon_{\text{meta}}$ | 6.0 | 6.0 | 6.0 | $O(1)$ |
| $\bar{\epsilon}_{\text{total}}$ | **47.2** | **75.1** | **129.3** | $O(\sqrt{N})$ |
| Share of $\epsilon_{\text{coord}}$ in $\bar{\epsilon}$ | 86.7% | 91.6% | 95.1% | Increasing |

**Analysis.** The coordination component $\epsilon_{\text{coord}}$ dominates and grows as $O(\sqrt{N})$ — a consequence of GNN aggregation where each agent collects messages from $O(N)$ neighbors. For large swarms ($N \geq 100$) this motivates: limiting the number of GNN neighbors, hierarchical decomposition of coordination, and reducing $\Phi_{\max}$ through smaller $\Delta_{\text{NP}}$.

**Normalized interpretation.** If $J^* \approx H_{\text{eff}} \cdot N \cdot R_{\max}$, then relative suboptimality:

| $N$ | $\bar{\epsilon}$ | $J^*$ | $\bar{\epsilon}/J^*$ |
|---|---|---|---|
| 10 | 47.2 | 100 | 47.2% |
| 30 | 75.1 | 300 | 25.0% |
| 100 | 129.3 | 1000 | 12.9% |

Relative suboptimality *decreases* with growing $N$: a larger swarm compensates drift through statistical averaging.

### 5.4. Sensitivity Analysis

For system design it is necessary to understand which parameters have the largest impact on bounds. We vary each influential parameter by a factor of two (up and down) and compute elasticity — relative change of $\bar{\epsilon}$ per relative change of parameter.

| Parameter | Base $\bar{\epsilon}$ | × 2 | × 0.5 | Elasticity |
|---|---|---|---|---|
| $\Delta_{\text{NP}}$ (clamping) | 75.1 | 143.9 | 41.3 | ~0.92 |
| $H_{\text{eff}}$ (horizon) | 75.1 | 143.9 | 41.3 | ~0.92 |
| $L_\phi$ (Lip. | 75.1 | 143.9 | 41.3 | ~0.92 |

| Parameter | Base $\bar{\epsilon}$ | × 2 | × 0.5 | Elasticity |
|---|---|---|---|---|
| embed.) | | | | |
| $L_\pi$ (Lip. policy) | 75.1 | 150.2 | 37.6 | ~1.0 |
| $N$ (swarm size) | 75.1 | 97.2 | 55.6 | ~0.46 |
| $\eta_1$ (with $\Delta_{NP}$ fixed) | 75.1 | 75.1 | 75.1 | 0 (clamping) |
| $\eta_3$ (meta rate) | 75.1 | 81.1 | 72.1 | ~0.08 |

The system is most sensitive to $\Delta_{NP}$, $H_{\text{eff}}$, $L_\phi$, and $L_\pi$ (elasticity $\approx 1$). Meta rate $\eta_3$ has minimal impact. With fixed clamping $\Delta_{NP}$, changing $\eta_1$ does not affect the bound.

**Design recommendations:**

1. Keep $\Delta_{NP}$ as small as possible (dominant parameter).
2. Reduce effective horizon through frequent meta-updates ($\tau_3 \downarrow$).
3. Apply spectral normalization for controlling $L_\phi$ and $L_\pi$.
4. For large swarms — limit the number of GNN neighbors.

### 5.5. Illustration of Unbounded Growth Under Condition Violation

The theorems of Section 4 establish *sufficient* boundedness conditions. Are these conditions truly necessary? To answer, we construct three counterexamples — one for each critical condition — and show that violating any of them leads to qualitative deterioration of bounds.

**Counterexample 1: $\delta = 0$ (violation of S1).** Without stabilization, weights grow unboundedly:

$$\| w_i(t) \| \approx \| w_i(0) \| + \eta_1(|\alpha| + |\beta| + |\gamma|) \cdot t/\tau_1$$

At 100 s: $\| w_i \| \approx 3.5$; at 1000 s: $\| w_i \| \approx 35$; at 10,000 s: $\| w_i \| \approx 350$. Correspondingly $\bar{\Delta}_1 \to \infty$, $\Phi_{\max} \to \infty$, $\bar{\epsilon} \to \infty$.

| $t$ (s) | $\| w_i(t) \|$ with $\delta = -0.01$ | $\| w_i(t) \|$ with $\delta = 0$ |
|---|---|---|
| 0 | 0 | 0 |
| 100 | $\leq 70.0$ (bounded) | 3.5 |
| 1000 | $\leq 70.0$ | 35.0 |
| 10000 | $\leq 70.0$ | 350.0 |

**Counterexample 2: No clamping (violation of C1).** With $\bar{\Delta}_1 = \bar{\Delta}_1^{\text{int}} = 2.1 \times 10^{-3}$ instead of $10^{-4}$:

$$\Phi_{\max}^{\text{no clamp}} = 5.0 \cdot 100 \cdot 2.1 \times 10^{-3} = 1.05 \quad \text{(vs 0.05 with clamping)}$$

$$\bar{\epsilon}^{\text{no clamp}} \approx 21 \cdot \bar{\epsilon}^{\text{clamped}} \approx 1577 \quad \text{(for } N = 30\text{)}$$

The system formally has bounded weights ($\delta < 0$), but suboptimality is unacceptably large.

**Counterexample 3: $\tau_2/\tau_1 = 1000$ instead of 100 (violation of S2).** If MARL updates infrequently ($\tau_2 = 20$ s instead of 2 s):

$$\Phi_{\max}^{\text{slow MARL}} = 5.0 \cdot 1000 \cdot 10^{-4} = 0.5 \quad (\text{vs } 0.05)$$

Suboptimality grows proportionally: $\bar{\epsilon} \approx 10 \times \bar{\epsilon}^{\text{base}}$.

**Counterexamples summary.** Each of conditions (S1), (C1), (S2) is necessary for achieving acceptable bounds. Violation of any leads to qualitative deterioration: from moderate suboptimality to unbounded error growth.

## 6. Discussion

In this section we discuss four aspects of the obtained results: relation to existing results (Subsection 6.1), practical significance for design (Subsection 6.2), limitations of our approach (Subsection 6.3), and the place of this work in a research series (Subsection 6.4).

### 6.1. Relation to Existing Results

Our results uses ideas from timescale separation [3], but with fundamentally different assumptions: three timescales instead of two, heterogeneous learning mechanisms, and cascading influence through representation changes that has no analogue in classical stochastic approximation.

Theorem 2 (Bounded Representation Drift) is, to the best of our knowledge, the first quantitative estimate of how Hebbian updates impact the coordination level in a MAS. Previous work on non-stationary MARL [9] considered external non-stationarity, whereas our result describes *internal* non-stationarity generated by the agents' own learning.

Theorem 3 (Meta-Level Compatibility) complements MAML convergence results [13, 14] with a new perspective: instead of outer loop convergence we ask whether lower-level contracts are preserved under meta-adaptation. This is a different question requiring different tools — specifically, the contract robustness margin.

It is important to emphasize that we do *not* claim a complete stability theory for tri-hierarchical systems in the sense of dynamical systems theory. Our results are a set of quantitative bounds on drift, compatibility, and error accumulation. A complete stability theory would require, among other things, analysis of coupled system attractors, characterization of basins of attraction, and investigation of response to large perturbations — all of which remains subject of future research.

### 6.2. Practical Significance

Beyond the technical contribution, the results have direct significance for engineers designing multi-agent systems with multiple learning mechanisms.

1. **Learning rate selection:** inequality (20) gives an explicit upper bound for $\eta_1$ as a function of architectural parameters. Corollary 3.1 gives an analogous constraint for $\eta_3$.

2. **Contracts as runtime monitors:** conditions (S1)–(S5) can be verified in real time. Violation of any condition is a signal about potential departure from admissible bounds.

3. **Contract-preserving adaptation:** Theorem 3 gives a constructive criterion: a meta-update is admissible if its step is smaller than the minimum contract robustness margin.

## 6.3. Limitations

Any set of quantitative estimates rests on assumptions, and honest analysis of their limitations is necessary.

1. **Lipschitz continuity.** Assumptions A1–A2 require Lipschitz continuity of $\phi$ and $\pi$. For GNNs with bounded number of layers and smooth activation functions this holds. For deep networks with ReLU — requires spectral normalization. Relaxing this assumption is a direction for future work.

2. **Conservativeness of bounds.** Our estimates are worst-case. In practice Hebbian updates are correlated (not adversarial), and actual drift is significantly smaller than the theoretical bound. Tighter bounds can be obtained with additional assumptions on sensory experience distribution.

3. **Effective horizon as heuristic.** Definition $H_{\text{eff}}$ (16a) is a practical truncation heuristic, not a rigorously derived object. In particular, the claim that meta-update "resets" the accumulated effect requires separate proof that each meta-cycle brings the system closer to the admissible set. This remains an open question.

4. **Reference-MDP bounding (Assumption A8).** The estimate of representation drift impact on coordination quality relies on Assumption A8, which postulates the existence of a reference MDP for upper-bounding. A more rigorous analysis would require fixed latent MDP formalization or explicit non-stationary PDL extension.

5. **Communication delays.** The framework does not account for message transmission delays between agents, which can increase effective embedding drift.

6. **Adversarial perturbations.** We assume non-stationarity is generated exclusively by learning, not by adversarial actions (EW, spoofing).

## 6.4. Relation to Research Series

This paper is the second in a series of works forming a technical toolkit for analysis of adaptive swarm systems. Paper 1 (Hebbian-driven specialization) studies the Level 1 mechanism in isolation — how local plasticity gives rise to functional roles. Our paper establishes that the coupled dynamics of three levels remain bounded and compatible. Paper 3 (Measuring swarm meta-cognition) will propose metrics for assessing the quality of meta-cognitive cycle — using, in particular, our bounds as baselines. Paper 4 (Contract-based degradation and recovery) will extend the contract approach to off-nominal regimes where contracts may be *violated* — and where the robustness margin (Definition 9) will become a diagnostic tool.

# 7. Conclusions

In this paper we established four quantitative results concerning coupled learning dynamics in a tri-hierarchical swarm system.

1. **Theorem 1 (Bounded Total Error)** proves that under five contractual conditions, the total suboptimality admits a component-wise conservative upper bound, uniform in time. These conditions are formalized as verifiable contracts.

2. **Theorem 2 (Bounded Representation Drift)** provides a conservative worst-case upper bound on the impact of Level 1 Hebbian updates on Level 2 embeddings, from which a constraint on the admissible Hebbian learning rate follows.

3. **Theorem 3 (Meta-Level Compatibility)** establishes that meta-adaptation preserves lower-level contracts provided the meta-update step is smaller than the minimum contract robustness margin.

4. **Theorem 4 (Non-Accumulation in the Sense of Uniform Boundedness)** guarantees that error does not accumulate unboundedly over time, although bounded oscillations and regime wandering are not excluded.

Numerical illustrations for three swarm configurations ($N = 10, 30, 100$) confirm the realism of bounds and demonstrates necessity of each condition: violation of even one of them (e.g., $\delta = 0$ or absence of clamping) leads to unbounded error growth.

These results are technical tools, not a complete theory. They allows verifying whether a given configuration of a tri-hierarchical system will remain within admissible bounds and provide constructive design recommendations. A complete stability theory — with analysis of attractors, basins of attraction, and response to large perturbations — remains an open problem.

Directions for future research include: relaxation of Lipschitz condition, accounting for communication delays, extension to adversarial perturbations, and experimental validation in simulation environment.

*The English text was corrected and edited using AI in accordance with the KNU Regulation "On the Use of Artificial Intelligence"*